\documentclass[runningheads]{llncs}

\usepackage{eccv}

\usepackage{eccvabbrv}

\usepackage{graphicx}
\usepackage{booktabs}

\usepackage{algorithm}
\usepackage{algorithmicx}
\usepackage{makecell}
\usepackage{amssymb}
\usepackage{booktabs}
\usepackage{colortbl}
\usepackage{xcolor}
\usepackage{amsmath}
\usepackage{multirow}
\usepackage{graphicx}
\usepackage{adjustbox}
\usepackage{url}
\usepackage{graphicx}
\usepackage{booktabs}
\usepackage{amsmath}
\usepackage{amssymb}
\usepackage{algorithm}
\usepackage{algorithmicx}
\usepackage{makecell}
\usepackage{amssymb}
\usepackage{booktabs}
\usepackage{colortbl}
\usepackage{amsmath}
\usepackage{multirow}
\usepackage{graphicx}
\usepackage{adjustbox}
\usepackage{tikz}

\usepackage{algpseudocode} % For additional math symbols
\usepackage{bm}       % For bold math symbols

\usepackage{colortbl}
\usepackage{booktabs} % For nicer tables
\usepackage{amsmath}
\usepackage{multirow}
\usepackage[utf8]{inputenc} % allow utf-8 input
\usepackage[T1]{fontenc}
\usepackage{nicefrac}       % compact symbols for 1/2, etc.
\usepackage{microtype}      % microtypography
\usepackage[dvipsnames]{xcolor}         % colors
\usepackage{graphicx}
\usepackage{subcaption}
\usepackage{colortbl}
\usepackage{makecell}
\usepackage{algorithm}
\usepackage{algpseudocode}
\usepackage{tcolorbox}
\usepackage{wrapfig}
\usepackage{footmisc}
\tcbuselibrary{listingsutf8}
\usepackage[dvipsnames]{xcolor}

\usepackage[cjk]{kotex}
\newcommand{\hs}[1]{#1}

\newcommand\ours{\texttt{Aether}\xspace}
\usepackage[accsupp]{axessibility}  % Improves PDF readability for those with disabilities.

\usepackage[pagebackref,breaklinks,colorlinks,citecolor=eccvblue]{hyperref}
\usepackage{orcidlink}

\begin{document}

\title{Isotropic Embedding Perturbations for\\ Robust Vision Language Encoders}
%Going Beyond Saturated Data Augmentations with Isotropic Embedding Perturbations

\titlerunning{Isotropic Embedding Perturbations}

\author{Hyesong Choi\inst{1,3}\thanks{This work was carried out during Hyesong Choi's internship at NAVER AI Lab.} \and
Daeun Kim\inst{2} \and
Song Park\inst{3} \and
Taekyung Kim\inst{3} \and
Byeongho Heo\inst{3} \and
Sangdoo Yun\inst{3} \and
Dongbo Min\inst{2}$^{\dagger}$ \and
Dongyoon Han\inst{3}\thanks{Corresponding authors.}}

\authorrunning{H. Choi et al.}

\institute{\textsuperscript{1}Soongsil Univ. \quad
\textsuperscript{2}Ewha W. Univ.\quad
\textsuperscript{3}NAVER AI Lab}

\maketitle

\begin{abstract}
Data augmentation is fundamental to training modern deep vision \hs{and multimodal} models. While individual methods, such as RandAug, CutMix, Mixup, RandErase, and DropPath, offer strong regularization effects, their combined use has saturated in performance due to overlapping functionalities, and aggressive pixel-level manipulations may disrupt delicate cross-modal alignment. This saturation motivates the search for a new augmentation axis within the embedding space rather than the input space. 
We introduce \ours, a simple plug-in method that applies \hs{diffusion-style} random perturbations in the embedding space via controlled alpha-mixing, \hs{specifically designed to provide isotropic regularization that remains semantically consistent}.
\hs{Inspired by feature-space perturbations in language models and image degradation in generative pretraining, \ours induces mild yet effective perturbations that smooth the representations without compromising the fine-grained structural information required for strong vision-language encoders.}
Across diverse architectures and across multiple recognition tasks, \ours delivers consistent gains over the advanced recipe combining CutMix, Mixup, DropPath, and RandAug—a level of improvement rarely observed with modern augmentation alternatives. \hs{Notably, \ours demonstrates superior effectiveness in multi-modal alignment, succeeding where traditional pixel-space augmentations fail by providing a stable, isotropic regularization signal that respects the integrity of the high-dimensional feature space.} Code is available at \url{https://github.com/naver-ai/aether}.
\end{abstract}

\begin{figure*}[t]
\centering
\small
\begin{subfigure}{0.38\textwidth}
    \centering
    \includegraphics[width=\textwidth]{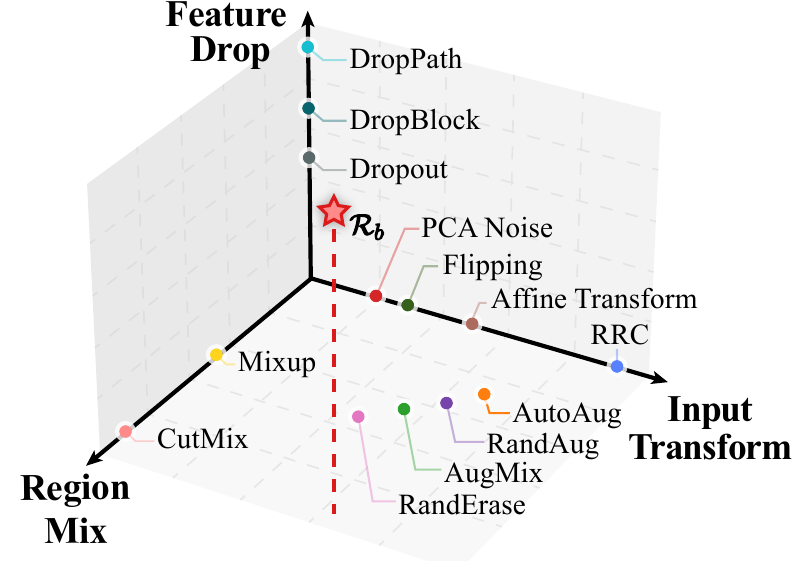}
    \caption{Illustrative augmentation space along grouped axes, with $\mathcal{R}_b$ = CutMix + Mixup + DropPath + RandAug}
\end{subfigure}%
\hfill
\begin{subfigure}{0.58\textwidth}
    \centering
    \includegraphics[width=\textwidth]{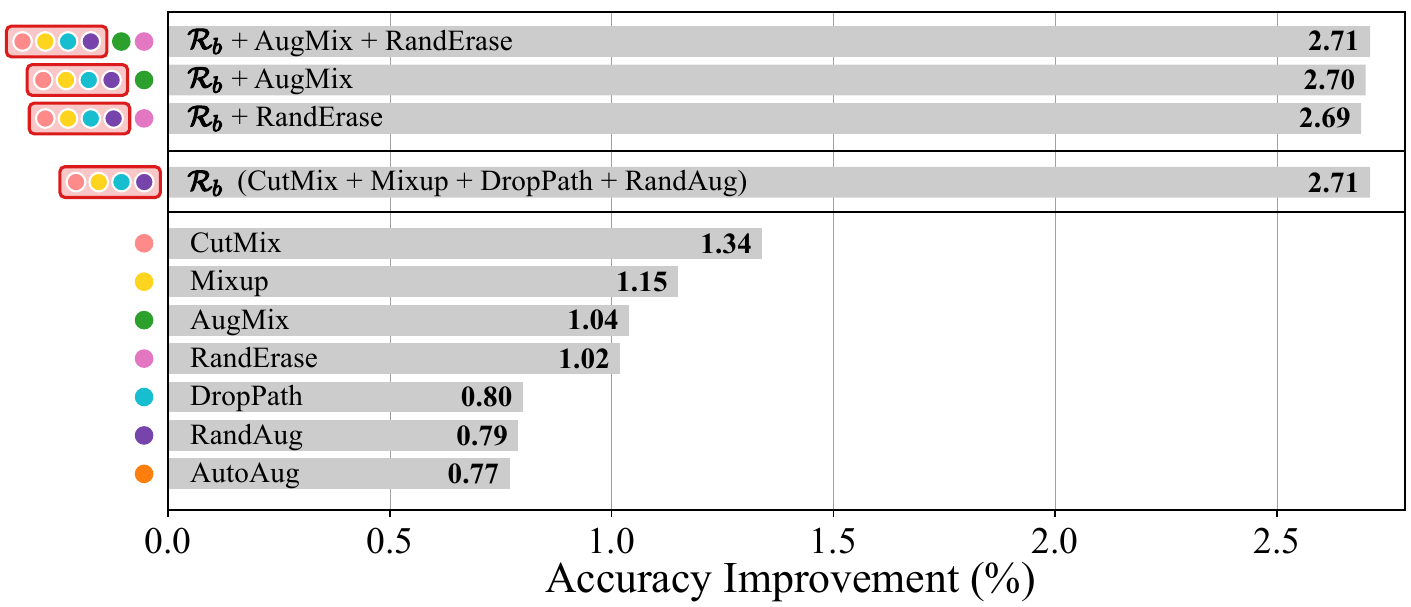}
     \caption{$\mathcal{R}_b$ appears saturated --  integrating other augmentations offers negligible gains}
\end{subfigure}%
% \vspace{-.5em}
\caption{\textbf{Where do we stand? Data augmentation may reach its limits in the era of robust \hs{visual and vision language encoders.}} When more is no more, we are now at \textit{the wall of augmentations} that limits further progress. Modern training pipelines (\eg, on ImageNet) are now saturated, where the standard recipe $\mathcal{R}_b$ dominates competing alternatives. Specifically, \textbf{(a)} conceptually, augmentation space can be drawn in three explicit axes—\emph{input transforms} (pixel/geometric manipulation in the image space), \emph{region-level mixing} (inter-sample blending within the pixel space), \emph{feature-level dropping} (sparsity regularization via setting paths to zero). \textbf{(b)} ImageNet accuracy gains vanish beyond the de facto recipe $\mathcal{R}_b$ = CutMix + Mixup + DropPath + RandAug. We argue that most modern augmentations collapse onto these three axes, resulting in highly overlapping regularization effects. Furthermore, we believe this collapse may also occur in training vision language models, as they also incorporate vision encoders. } %\hs{Furthermore, they often fail to provide complementary regularization for vision language encoders; their pixel-level manipulations disrupt delicate cross-modal alignment and, after the stem layer, manifest as anisotropic noise that fails to effectively regularize the high-dimensional feature space.}}
\label{fig:augs_no_complementary}
\vspace{-.5em}
\end{figure*}
% axis span

\section{Introduction}
Data augmentation is a staple for the generalization capability of vision models. It has first delivered large gains in Convolutional Neural Networks (CNNs)~\cite{vgg,googlenet,resnet,tan2019efficientnet} by enriching data diversity and mitigating overfitting. After the emergence of Vision Transformers (ViT)~\cite{vit,deit,touvron2022deit}, %and Transformer-based training~\cite{mae, xie2022simmim, diffmae,maskdit,diffmim}, 
they have made augmentation even more crucial; this is likely because ViTs have a higher capacity relying on global self-attention, while lacking spatial inductive biases, which can make them sensitive to small perturbations. In downstream fine-tuning with ViTs \hs{and vision language alignment in vision language models (VLMs)}, this issue becomes central: effective transfer of pre-trained parameters depends on carefully chosen augmentations that inject task-relevant inductive biases and curb overfitting under limited labels. 

Tremendous efforts have been made to discover a golden augmentation recipe; however, we now face \textit{the wall of the standard recipe} combining CutMix~\cite{yun2019cutmix}, Mixup~\cite{zhang2017mixup}, DropPath~\cite{huang2016deep}, and RandAug~\cite{cubuk2020randaugment} (we denote it as $\mathcal{R}_b$) for fine-tuning ViTs (often combined with Label Smoothing~\cite{Inceptionv3} 
and Dropout~\cite{srivastava2014dropout}). %; we denote it by $\mathcal{R}_b$. 
In practice, despite broad adoption and strong results, its gains have reached a performance plateau: many prior works~\cite{tan2019efficientnet, deit, han2021rethinking, wightman2021resnet, steiner2021train, touvron2022deit, dehghani2023scaling, kim2024densenets, heo2025maksub} employed or slightly extended the standard setup, yet the combination remains the default and has plateaued in augmentation diversity. 

Fig.~\ref{fig:augs_no_complementary}(a) illustrates a conceptual augmentation space, where most methods lie within three dominant axes---input transforms, region-level mixing, and feature-level dropping. Many methods remain largely confined to the input space and region mixing, relying on photometric/geometric transforms and inter-sample mixing. \hs{While effective for uni-modal vision, these pixel-level manipulations often fail to provide complementary regularization for vision language encoders~\cite{radford2021learning,fini2025multimodal,zhai2023sigmoid,tschannen2025siglip}, in which the aggressive mixing of images may yield diminishing returns for the delicate fine-grained alignment between text and images.} In the feature space, regularization has been restricted to passive feature dropping (\eg, DropPath~\cite{huang2016deep}, Dropout~\cite{srivastava2014dropout}). This limitation in advancement causes augmentations to cluster closely along these three axes, leaving other valuable directions in the feature space underexplored. Indeed, Fig.~\ref{fig:augs_no_complementary} (b) displays that stacking additional augmentations on top of $\mathcal{R}_b$ yields almost no improvements. %\textit{Why does the standard recipe saturate?} 
We argue that this saturation stems from the significant conceptual overlap among existing methods, \hs{which not only operate along limited axes but also manifest as anisotropic noise after passing through the stem layer, failing to effectively regularize the high-dimensional feature space.}
%which predominantly operate along limited axes. 

\textit{Can we discover a new method that operates on top of the standard recipe?}
We draw inspiration from two trends in modern representation learning. First, \textbf{Feature-Space perturbations}: while standard vision pipelines heavily rely on spatial and photometric manipulations~\cite{cubuk2020randaugment, yun2019cutmix, zhang2017mixup}, using stochastic perturbations in the \textit{embedding space} has been highly effective in other modalities, notably in language modeling~\cite{hua2021noise, nukrai2022text, jain2024neftune}. Second, \textbf{Image degradation}: generative pretraining~\cite{mae,ho2020denoising, diffmae} demonstrated the power of \textit{intentional data degradation} and subsequent recovery as a mechanism for learning robust, generalized features. \textit{What if degradation-based regularization were realized as an embedding-space perturbation?} 

Building upon these observations, we propose \ours\footnote{The name evokes the image of a perturbed signal softly permeating the representation space, much like air.}, a simple embedding-space data augmentation method for improved representation learning.
\ours\ employs a \hs{diffusion-style} controlled alpha-mixing mechanism to smoothly mix input embeddings with an isotropic Gaussian. \hs{By injecting isotropic perturbations directly into the embedding space, \ours\ provides a new axis of augmentation that enhances the generalization of vision language encoders without destroying the underlying semantic structure critical for cross-modal alignment.} This allows for controlled perturbations in the feature space without abruptly corrupting semantic information, effectively smoothing the representation space.
We argue \ours\ complements the standard setup $\mathcal{R}_b$ in a distinct and complementary manner and integrates effectively as a plug-in. Furthermore, our analysis reveals that (1) \ours\ enhances localization by improving attention capabilities, presumably because smoother embeddings improve sensitivity to fine details; (2) \ours\ further acts as an embedding-level regularization due to the embedding smoothing process, thus leading to improved robustness; (3) \ours\ demonstrates a strong ability to learn more generalized representations, as revealed by its flatter loss landscape.%셋이 비슷한 느낌으로 보여서 analysis 수정 후 조금더 특색있어보이게 writing 수정 예정

We conduct extensive experiments with \ours\ across diverse architectures, including ViT-{S,B,L}~\cite{vit}, multi-modal models (CLIP~\cite{radford2021learning}, AIMv2~\cite{fini2025multimodal}, SigLip 2~\cite{tschannen2025siglip}), and modern self-supervised frameworks, masked image modeling (MIM)~\cite{mae, xie2022simmim} and diffusion-MIM~\cite{diffmae, maskdit, diffmim}). When combined with standard augmentations, \ours\ yields up to a 3.49\%p performance gain. Beyond ViTs, consistent improvements on CNNs (ResNet-26/50) and hierarchical transformers (Swin V2-L~\cite{liu2022swin}) indicate that its effect is architecture-agnostic. We further evaluate \ours\ on a range of recognition tasks---image classification, fine-grained visual classification (FGVC), semantic segmentation, object detection, and instance segmentation. Tasks such as FGVC relying on subtle, part-level cues benefit most: \ours\ enhances fine-detail sensitivity by mitigating spurious localized attention and promoting a broader, more coherent attention distribution, yielding consistent gains on fine-grained benchmarks~\cite{cub, nabirds}.

\vspace{-0.1em}
\section{Related Work}
\noindent\textbf{Data augmentation} has been a cornerstone for improving the generalization of deep vision models. The strategies have evolved from basic input-space manipulations, such as geometric and photometric transformations~\cite{krizhevsky2012imagenet}, to sophisticated policy-based methods like AutoAug~\cite{cubuk2019autoaugment} and RandAug~\cite{cubuk2020randaugment}. Concurrently, powerful mixing strategies emerged, including Mixup~\cite{zhang2017mixup} and CutMix~\cite{yun2019cutmix}, which combine images via patch pasting. These methods, often used alongside architectural regularization like DropPath~\cite{huang2016deep}, have coalesced into a de facto standard recipe ($\mathcal{R}_b$) widely adopted in high-performance vision training pipelines~\cite{rw2019timm, wolf2020transformers}. While highly effective, this standard recipe has reached saturation. Recent studies~\cite{devries2017improved, zhang2021understanding} and our analysis (Fig.~\ref{fig:augs_no_complementary}a) indicate that stacking these strong augmentations yields diminishing returns. \hs{Critically, for vision language encoders, these aggressive pixel-mixing strategies often disrupt the delicate fine-grained alignment between visual and textual modalities, making them largely ineffective for multi-modal tasks.} This saturation \hs{and ineffectiveness} suggest that current methods, which predominantly manipulate the \textit{input space}, \hs{not only overlap in their regularization effects but also fail to provide an isotropic regularization signal for high-dimensional representations.}\\

%largely overlap in their regularization effects, motivating the exploration of distinct axes of regularization.

\noindent\textbf{Embedding-/feature-space augmentation.} While input-space augmentation has saturated in vision, regularization within the \textit{embedding space} remains largely underexplored. In contrast, perturbing embeddings via stochastic mechanisms is an established technique in Natural Language Processing (NLP) to enhance generalization by smoothing the representation space~\cite{miyato2017adversarial, hua2021noise, jain2024neftune}. Furthermore, recent vision SSL methods~\cite{mae, ho2020denoising} demonstrate the power of intentionally degrading and recovering features for representation learning. \ours\ synthesizes these insights \hs{by introducing a diffusion-style controlled alpha-mixing mechanism. By injecting isotropic perturbations directly at the embedding level, \ours\ introduces a new regularization method that overcomes the limitations of input-space methods, which manifest as anisotropic noise after passing through the stem layer and thus fail to regularize the feature space effectively. Unlike previous approaches that interfere with cross-modal correspondence, \ours\ preserves the integrity of the semantic structure, making it uniquely suited for robust vision language alignment.}
%addressing the saturation of input-space methods by introducing a controlled embedding smoothing strategy designed to provide a distinct and complementary regularization signal.

\section{Method}
This section begins by presenting the background of our method, focusing on prior methods of embedding-space augmentations and image degradations. We then detail the mechanics of \ours, motivated by the insight that current approaches are not seamlessly compatible with smoothing-oriented augmentation.

\subsection{Background}
\noindent\textbf{Embedding-space augmentations in language modeling.} In language modeling, embeddings are often perturbed or masked to enhance generalization~\cite{bert,miyato2017adversarial,zhu2020freelb}. Among various perturbation strategies, stochastic perturbation (\eg, using Gaussian noise) has been commonly employed for this purpose~\cite{hua2021noise, nukrai2022text, jain2024neftune}. Specifically, given token embeddings $\mathbf{e}_i \in \mathbb{R}^d$, a smooth perturbation is applied like $
\tilde{\mathbf{e}}_i = \mathbf{e}_i + \boldsymbol{\epsilon}_i, \text{\ s.t.\ } \boldsymbol{\epsilon}_i \sim \mathcal{N}(\mathbf{0}, \sigma^2 \mathbf{I})$. Intuitively, this encourages invariance to small perturbations in embeddings, leading to more robust representations. Our method draws inspiration from this, applying stochastic perturbations to tokens, rather than to the image input, as an additional exploration of the visual augmentation space.

\noindent\textbf{Image degradation in generative pretraining.} Modern pretraining frameworks~\cite{mae, xie2022simmim, maskdit, diffmae, diffmim, ho2020denoising, song2020denoising, nichol2021improved, rombach2022high, dalle, imagen, choi2024emerging, choi2024salience, son2024sg}, particularly in self-supervised and generative modeling, rely heavily on the principle of intentional image degradation for reconstruction. For example, methods like Masked Autoencoders (MAE)~\cite{mae} employ aggressive masking, while Denoising Diffusion Models (DDM)~\cite{ho2020denoising, song2020denoising, nichol2021improved, rombach2022high, dalle, imagen} utilize a gradual degradation process. These frameworks are believed to learn robust features by training the network to reconstruct the original \textit{clean image} from a \textit{degraded one}, and may provide further insights for their use as augmentations.

%넵 이건 제가 수정할게요. 표 형태도 좋고 좋습니다.

\subsection{Introducing the Proposed Method}

\noindent\textbf{Motivation and pilot study.} Our inspiration stems from two key observations: (1) data augmentation methods in vision have underexplored feature-space regularizations \hs{that are crucial for maintaining cross-modal integrity}, and (2) \hs{diffusion-style} image degradation–reconstruction approaches in self-supervised learning have proven highly effective. Bridging these perspectives may lead us to breakthroughs beyond the current performance saturation. To this end, we revisit the effectiveness of perturbations in the embedding space for training \hs{both vision and vision language models}; we further argue that applying such perturbations in the input space (\eg, perturbation-based photometric distortions or  spatial transformations) may suffer from inherent limitations:
\begin{enumerate}
    
    \item \textit{Input-level perturbations \hs{disrupt the delicate cross-modal alignment and fine-grained structures}}.
    %may disrupt low-level spatial coherence and fine-grained structures}.
    Before any encoding, \hs{pixel intensities represent textures and shapes essential for semantic alignment; aggressive manipulations like CutMix~\cite{yun2019cutmix} and Mixup~\cite{zhang2017mixup} lose these subtle details, interfering with the precise correspondence between visual tokens and textual descriptions required for vision encoders.}
    %pixel intensities vary to represent texture and shape, but the perturbation loses these subtle details.

    \item \textit{Perturbing pixels does not translate uniformly into the embedding space, diminishing \hs{regularization impact}.}
    %its impact.}
    After passing through the stem layer (\eg, patchification or initial convolutions), isotropic pixel perturbations (\ie, uniformly spread across all dimensions) become \textit{anisotropic} in the embedding space, thereby reducing the effectiveness of pixel-level perturbation methods, as consistently observed in prior works.
\end{enumerate}
% \vspace{0.3em}
%Before going further, %to validate our arguments, our pilot study reveals this; the effect of input-level perturbations diminishes and barely contributes to the performance (see \cref{tab:pilot_study_level}). Furthermore, we measure the token-perturbation covariance
%spectrum (eigenvalue ratio $\lambda_{\max}/\lambda_{\min}$) over 50K ImageNet-val images,together with CUB accuracy where part-level cues are most anisotropy-sensitive (Tab.~\ref{tab:anisotropy}). Pixel-level noise reaches a ratio of $8.4$, empirically confirming that the patch-embedding operator $\mathcal{P}$ collapses isotropic pixel noise onto a few dominant directions, whereas embedding-level injection stays near-isotropic($\approx 1.15$). This anisotropy disproportionately harms fine-grained cues, explaining why the CUB gap exceeds the ImageNet-1K gap.\\
Before proceeding, we conduct a pilot study comparing input-level and embedding-level perturbations to quickly test our claims. We primarily evaluate fine-grained recognition using CUB~\cite{cub} accuracy, while also reporting ImageNet-1K~\cite{imagenet} accuracy as a preliminary reference point for the general image understanding results presented later. In addition, we measure the token-perturbation covariance spectrum on the ImageNet validation set using the eigenvalue ratio $\lambda_{\max}/\lambda_{\min}$ to quantify the anisotropy induced by each perturbation type.

\cref{tab:pilot_study_level} demonstrates that the pilot study results support our claim: embedding-level perturbations achieve better CUB performance, which suggests that they more effectively preserve and regularize fine-grained visual cues. Moreover, pixel-level perturbations yield a high anisotropy ratio of $8.4$; by contrast, embedding-level perturbations remain nearly isotropic, with a ratio of approximately $1.15$.  This anisotropy disproportionately affects fine-grained cues, explaining why the CUB gap is larger than the ImageNet-1K gap.

%\begin{figure}[h]
%\centering
%\includegraphics[width=1\linewidth, trim={0 200 0 0},clip]{images/embed_vs_img.png}
%\vspace{-2em}
%\caption{\textbf{Embedding- vs. pixel-level perturbations.} Different capabilities in capturing fine-grained structures.} 
%\label{fig:pilot_study_fig}
%\vspace{-1em}
%\end{figure}

% \begin{table}[t]
% \centering
% \addtolength{\tabcolsep}{3pt}
% \small
% \caption{\textbf{Embedding- vs. pixel-level perturbations.} The effect of input perturbation diminishes, resulting in minimal gains.}
% \label{tab:pilot_study_level}
% % \vspace{-1em}
% \begin{tabular}{l ccc}
% \toprule
% \textbf{Metric} & \textbf{Baseline} & \textbf{Embed.} & \textbf{Pixel}   \\
% \midrule
% ImageNet-1K Top-1 (\%) & 79.02 & \textbf{82.25} & 80.17  \\
% \bottomrule
% \end{tabular}
% \vspace{-.5em}
% \end{table}

% \begin{table}[t]
% \centering
% \tabcolsep=1em
% \caption{Empirical anisotropy of pixel- vs.\ embedding-level perturbations, measured as
% the token-covariance ratio $\lambda_{\max}/\lambda_{\min}$ over 50K ImageNet-val images.}
% \label{tab:anisotropy}
% \begin{tabular}{lcc}
% \toprule
%  & $\lambda_{\max}/\lambda_{\min}$ & CUB Acc \\
% \midrule
% Pixel-level          & 8.4\,$\pm$\,0.5  & 79.33\,$\pm$\,0.11 \\
% Embed-level (Aether) & 1.2\,$\pm$\,0.05 & 80.89\,$\pm$\,0.09 \\
% \bottomrule
% \end{tabular}
% \end{table}

\begin{table}[h]
\vspace{-2em}
\centering
\tabcolsep=1em
\caption{\textbf{Embedding- vs. pixel-level perturbations.} The effect of input perturbation diminishes, resulting in minimal gains. Specifically, general understanding performance, fine-tuned performance measured by CUB accuracy, and empirical anisotropy consistently reveal which perturbation type is more effective, supporting our claim. }
\label{tab:pilot_study_level}
\begin{tabular}{lccc}
\toprule
 & $\lambda_{\max}/\lambda_{\min}$ & CUB (\%) & ImageNet-1K (\%) \\
\midrule
Pixel-level          & 8.4\,$\pm$\,0.5  & 79.33\,$\pm$\,0.11 &  80.17 \\
Embed-level          & \textbf{1.15\,$\pm$\,0.05} & \textbf{80.89\,$\pm$\,0.09} & \textbf{82.25}\\
\bottomrule
\end{tabular}
\vspace{-.5em}
\end{table}

%In contrast, operating directly within the embedding space allows for the introduction of truly isotropic perturbations. This ensures that the regularization signal is evenly distributed, smoothing the feature space precisely at the level where the model performs its computations, while preserving the spatial relationships maintained by token boundaries and positional structures.

\noindent\textbf{Design principle.} %The design of \ours\ is guided by the necessity of inducing meaningful regularization while rigorously maintaining semantic fidelity. 
While simple additive perturbations are utilized in other modalities~\cite{jain2024neftune}, directly adding stochastic elements to high-dimensional vision embeddings can abruptly shift the representations, potentially corrupting the underlying semantic structure or overpowering the original signal. Therefore, an effective mechanism must provide controlled, smooth perturbations that gently reshape the feature landscape rather than disrupt it. The regularization should encourage robustness to mild variations without compromising the essential information contained within the embeddings.

Among existing perturbation methods, \ours\ employs a simple yet controlled alpha-mixing strategy, \hs{directly inspired by the variance-preserving forward process at a specific timestep $t$ in denoising diffusion models~\cite{ho2020denoising, song2020denoising, nichol2021improved}.}
%inspired by the \textit{variance-preserving perturbation processes} in denoising diffusion models~\cite{ho2020denoising, song2020denoising, nichol2021improved} to realize these principles.
We smoothly blend the input embeddings with stochastic perturbations \hs{$\boldsymbol{\eta}_t$ drawn from an isotropic perturbation corresponding to timestep $t$}.
%drawn from an isotropic corruption.
Given an embedding tensor $\mathbf{Z} \in \mathbb{R}^{N\times d}$ (where $N$ is the number of tokens and $d$ is the embedding dimension), \ours\ generates a perturbed embedding $\tilde{\mathbf{Z}}$ as follows:
%\begin{equation}
%\tilde{\mathbf{Z}} = \sqrt{\alpha} \cdot \mathbf{Z} + \sqrt{1 - \alpha} \cdot \boldsymbol{\eta}, \quad \text{where } \boldsymbol{\eta} \sim \mathcal{N}(\mathbf{0}, \mathbf{I}).
%\label{eq:ours_blending}
%\end{equation}
\hs{\begin{equation}
\tilde{\mathbf{Z}} = \sqrt{\bar{\alpha}_t} \cdot \mathbf{Z} + \sqrt{1 - \bar{\alpha}_t} \cdot \boldsymbol{\eta}_t, \quad \text{where } \boldsymbol{\eta}_t \sim \mathcal{N}(\mathbf{0}, \mathbf{I}).
\label{eq:ours_blending}
\end{equation}}
\hs{The timestep $t$ acts as a discrete proxy for the augmentation intensity, where the hyperparameter $\bar{\alpha}_t$ follows a pre-defined noise schedule. This formulation ensures a smooth interpolation along the diffusion trajectory, between the original clean embedding \textbf{$t=0$} and a noise-shrouded representation at higher $t$ values.}
%The hyperparameter $\alpha \in (0, 1)$ stands for the blending ratio, directly controlling the magnitude of the perturbation. This formulation ensures a smooth interpolation between the original embedding ($\alpha{=}1$) and a pure stochasticity ($\alpha{=}0$).

Unlike naive additive methods, this \hs{variance-preserving} approach \hs{leverages the mathematical properties of diffusion forward-mapping to help preserve the overall feature statistics, preventing abrupt shifts in the distribution during training. By treating augmentation as a diffusion-based degradation process, we encourage the model to learn representations that are invariant to isotropic variations in the latent manifold. This could effectively smooth the representation landscape and provide a new axis of regularization signal that is complementary to the standard recipe $\mathcal{R}_b$. Unlike pixel-space perturbations, which may disrupt cross-modal correspondence, \ours preserves semantic integrity by injecting perturbations directly into the representation space where the model performs cross-modal alignment. This suggests that \ours may be better suited for robust vision-language representation learning. We further support this intuition through the theoretical analysis below.}\\

%helps preserve the overall magnitude of the embeddings, preventing abrupt shifts in the feature distribution during training. This controlled introduction of stochasticity encourages the model to learn representations that are invariant to mild variations in the feature space. This effectively smooths the representation landscape and provides a distinct regularization signal that is complementary to the standard recipe $\mathcal{R}_b$. \ours\ functions as a simple, plug-in module that can be applied at the input embedding layer or inserted between intermediate blocks of the architecture.

\noindent\textbf{Theoretical backup: why at the embedding space?}
% We argue that the location of the perturbation is critical: effective regularization requires \textit{isotropic} smoothing (uniformly spread across dimensions) in the space where computation occurs.
We provide an informal theoretical justification for our approach. Our conjecture is that isotropic perturbations in the embedding space, uniformly spread across all token dimensions, are more desirable, whereas pixel-level perturbations incur bias.
Specifically, let an input image be $x\in\mathbb{R}^{H W C}$ and assume the patch-embedding operator is expressed as a linear matrix, $\mathcal{P} \in \mathbb{R}^{N d \times H W C}$ (\eg, a stride-$p$ conv). Injecting pixel-level perturbation $\epsilon\!\sim\!\mathcal{N}(0,\sigma^2 I)$ leads to
\begin{equation}
z_0' \;=\; \mathcal{P}(x+\epsilon) \;=\; \mathcal{P}x \;+\; \mathcal{P}\epsilon, %\underbrace{\mathcal{P}(\epsilon}_{\text{embedding perturbation after patchify}},
\end{equation}
so the perturbation in the embedding space is $\epsilon_{\text{tok}}=\mathcal{P}\epsilon$ with covariance
\begin{equation}
\mathrm{Cov}[\epsilon_{\text{tok}}] \;=\; \sigma^2\,\mathcal{P}\mathcal{P}^\top .
\end{equation}
Thus, the covariance is no longer isotropic, resulting in a biased perturbation that concentrates disproportionately on certain directions or channels rather than being evenly distributed.
Moreover, pixel-space perturbation disrupts \emph{spatial alignment} before the stem layer (\eg, patchification or a set of convolutions), entangling perturbation across neighboring pixels inside each patch and degrading fine structure that is critical for fine-grained data like FGVC datasets.

In contrast, injecting perturbation \emph{in the embedding space} preserves isotropy, magnitude, and alignment at the level where the model actually computes:
\begin{equation}
z_0' \;=\; \mathcal{P}x \;+\; \eta_0,\qquad \eta_0\sim\mathcal{N}(0,\sigma_e^2 I_{N d}).
\end{equation}
Let $z_l$ be the $l$-th layer's embedding without perturbations and $z_l'$ be the result from $z_0'$. For a residual block $f_l(\cdot)$, the perturbation after {\small $(l+1)$}-th layer can be
\begin{equation}
    \eta_{l+1} = z_{l+1}' - z_{l+1} = z_{l}' - z_{l} + f_l(z_{l}') - f_l(z_{l}) = (I + J_{f_l}) \eta_{l},
\end{equation}
% \begin{equation}
% \begin{split}
% z_{l+1} \;=\; z_l \;+\; f_l(z_l), \quad\Rightarrow\quad
% \underbrace{\eta_{l+1}}_{\text{perturbation}} \\ \approx\; (I + J_{f_l}(z_l))\,\underbrace{\eta_l}_{\text{perturbation}},
% \end{split}
% \end{equation}
where $J_{f_l}$ denotes the discrepancy between residual-side feature computations with and without perturbations, which is expected to remain small in practice. We observe that the skip connection provides a \emph{shortcut} for the perturbation $\eta$ to propagate \emph{without attenuation by weights} like $\mathcal{P}$. 
This first-order effect of $\eta$ remains observable even in practice, for pre-normalization architectures. 
% With LayerNorm ($\mathrm{LN}$) centered and scale-learned, the first-order effect of $\eta$ remains observable in the attention logits:
% \begin{equation}
% A = \mathrm{softmax}\!\left(\frac{QK^\top}{\sqrt{d}}\right),
% \\
% Q{=}W_Q\,\mathrm{LN}(z_l), K{=}W_K\,\mathrm{LN}(z_l),
% \end{equation}
%Therefore, we believe that the resulting gradients explicitly encourage the network to learn smoother representations while preserving token boundaries and positional structure. 

Empirically, this yields (i) stronger early-layer diversity (signal present at the level where computation occurs) and (ii) late-layer recovery via residual aggregation—benefits that vanish when perturbation is injected in pixels and then suppressed by $\mathcal{P}$. We argue that propagation through layers encourages the learning of more generalized features, owing to residual perturbations that persist across layers.\\

\begin{figure}[t]
\centering
\includegraphics[width=.98\linewidth]{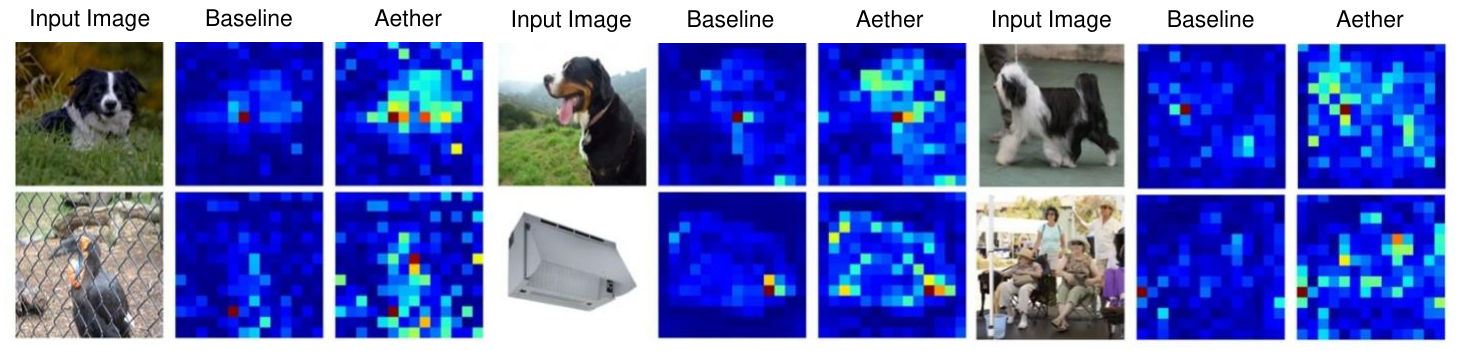}
% \vspace{-1em}
\caption{\textbf{\ours promotes broader and more coherent attention.}
Compared to the baseline trained with $\mathcal{R}_b$ (middle), which often exhibits imprecise attention patterns with spurious localized peaks, the \ours-augmented model (right) distributes attention more smoothly and consistently across foreground regions. This mitigates over-reliance on individual tokens and enhances semantic coherence.}
\label{fig:att_map}
\vspace{-.5em}
\end{figure}

\begin{figure}[b]
\centering
\vspace{-3em}
\begin{minipage}{0.9\linewidth}
\begin{algorithm}[H]
\footnotesize
\caption{PyTorch-like code for \ours}
\label{alg:ours}
\begin{algorithmic}[1]
\Require Blending factor $\alpha \in (0, 1)$
\Function{Add\ours}{$\mathbf{Z}, \alpha$}
    \State $s \gets \sqrt{\alpha}$,\quad $n \gets \sqrt{1-\alpha}$
    \State $\boldsymbol{\eta} \sim \mathcal{N}(\mathbf{0}, \mathbf{I})$ \Comment{\footnotesize same tensor shape as $\mathbf{Z}$}
    \State \Return $\mathbf{\tilde{Z}} \gets s \cdot \mathbf{Z} \;+\; n \cdot \boldsymbol{\eta}$
\EndFunction
\\
\State $x \gets \texttt{DataLoader}(\text{ImageNet},\ \text{augmentation}=\mathcal{R}_b)$
\State $\mathbf{Z} \gets \texttt{PatchEmbed}(x)\;+\;\texttt{PosEmbed}$ %\Comment{\footnotesize Input embeddings $\mathbf{Z}$}
\State \textcolor{RubineRed}{$\mathbf{Z} \gets \texttt{Add\ours}(\mathbf{Z}, \alpha)$} \Comment{\footnotesize Single added line for \ours}
\State $x \gets \texttt{Encoder}(\mathbf{Z})$
\end{algorithmic}
\end{algorithm}
\end{minipage}
\vspace{-.5em}
\end{figure}

\noindent\textbf{Implementation.} Algorithm~\ref{alg:ours} illustrates how \ours integrates into the standard fine-tuning pipelines~\cite{vit,deit,mae, touvron2022deit,heo2025maksub}. Given an existing training setup using strong augmentations, \ours requires only \textit{a single additional line}---injecting isotropic perturbations into the embeddings after patch embedding and positional encoding. This simplicity shows the modularity of our method: it acts as a lightweight, plug-and-play augmentation component that operates in the feature space without altering the model architecture or training procedure.

\section{Empirical Analyses}
\label{sec:whywork}
This section analyzes the mechanisms underlying \ours.\\

\noindent\textbf{Improved attention capability for localization.}
Fig.~\ref{fig:att_map} shows that the baseline trained with $\mathcal{R}_b$ exhibits narrow attention with spurious peaks on background regions or over-concentration on limited foreground areas. In contrast, \ours produces smoother attention maps with broader coverage over task-relevant regions. We attribute this behavior to embedding-space perturbations, which enhance representation robustness and encourage attention to previously weakly activated tokens. Pixel-level perturbations, by comparison, may diminish after patchification (\ie, large-kernel convolution), limiting their effect on token representations.\\

\begin{figure}[t]
\centering
\hspace{-1.5em}
\includegraphics[width=0.8\linewidth]{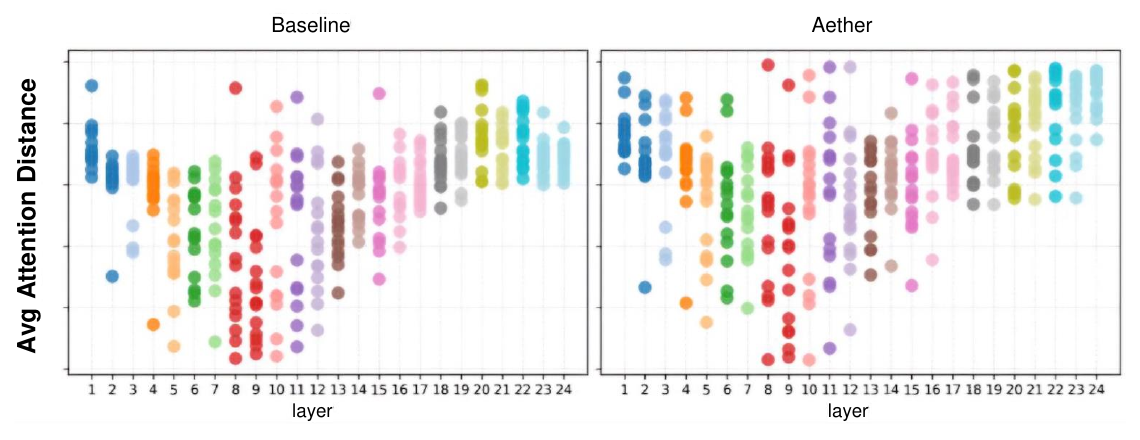}
\vspace{-.5em}
\caption{\textbf{\ours expands attention distance across layers.} We measure the average attention distance across transformer layers for ViT-L. The baseline model ($\mathcal{R}_b$) shows a tendency for narrowed attention in later layers. In contrast, \ours consistently maintains a broader attention distance, facilitating better integration of local cues with global context without architectural modifications.}
\label{fig:att_distance}
\vspace{-.5em}
\end{figure}

\noindent\textbf{Expanded attention distance.}
We measure the average attention distance across layers, representing the spatial range of token interactions. As shown in Fig.~\ref{fig:att_distance}, the baseline progressively narrows attention in deeper layers. \ours maintains significantly broader attention distances, enabling stronger coupling between local cues and global context throughout the network.\\

\begin{figure}[t]
\centering
\includegraphics[width=.7\linewidth]{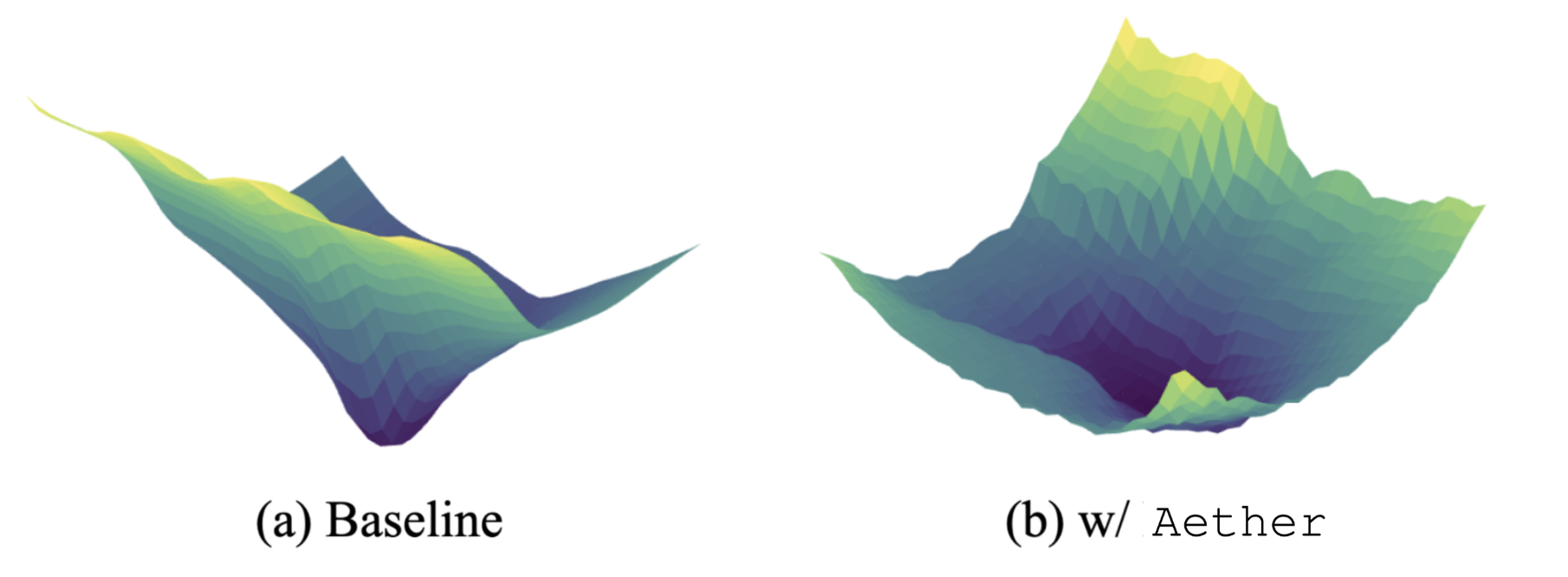}
\vspace{-1em}
\caption{\textbf{Loss visualization.} We plot the loss surfaces: (a) baseline + $\mathcal{R}_b$ vs. (b) baseline + $\mathcal{R}_b$+\ours. \ours converges to a flatter minimum, which suggests \ours learns more generalized features.} 
\label{fig:loss_flatness}
\vspace{-.5em}
\end{figure}

\noindent\textbf{Flattened loss landscape.} Loss landscape geometry provides a proxy for generalization. As visualized in Fig.~\ref{fig:loss_flatness}, the baseline converges to a sharp minimum, whereas \ours yields a substantially flatter basin, indicating improved robustness and stability.\\

%\begin{figure}[t]
%\centering
%\includegraphics[width=0.5\linewidth]{images/denoiser_wcleanimg2.png}
%\vspace{-2em}
%\caption{\textbf{Attention visualizations as applying 20$\times$ perturbations.}  Intriguingly, only \ours\ progressively suppresses perturbation—becoming \textit{increasingly similar to the clean image} -- as layers approach the output, yielding class-consistent latents; whereas, conventional augmentations such as CutMix, Mixup, and Dropout retain residual artifacts.}
%\label{fig:5_1_denoiser1}
%\vspace{-.5em}
%\end{figure}

\begin{figure*}[t]
\centering

\begin{subfigure}[t]{0.44\linewidth}
    \centering
    \includegraphics[width=\linewidth]{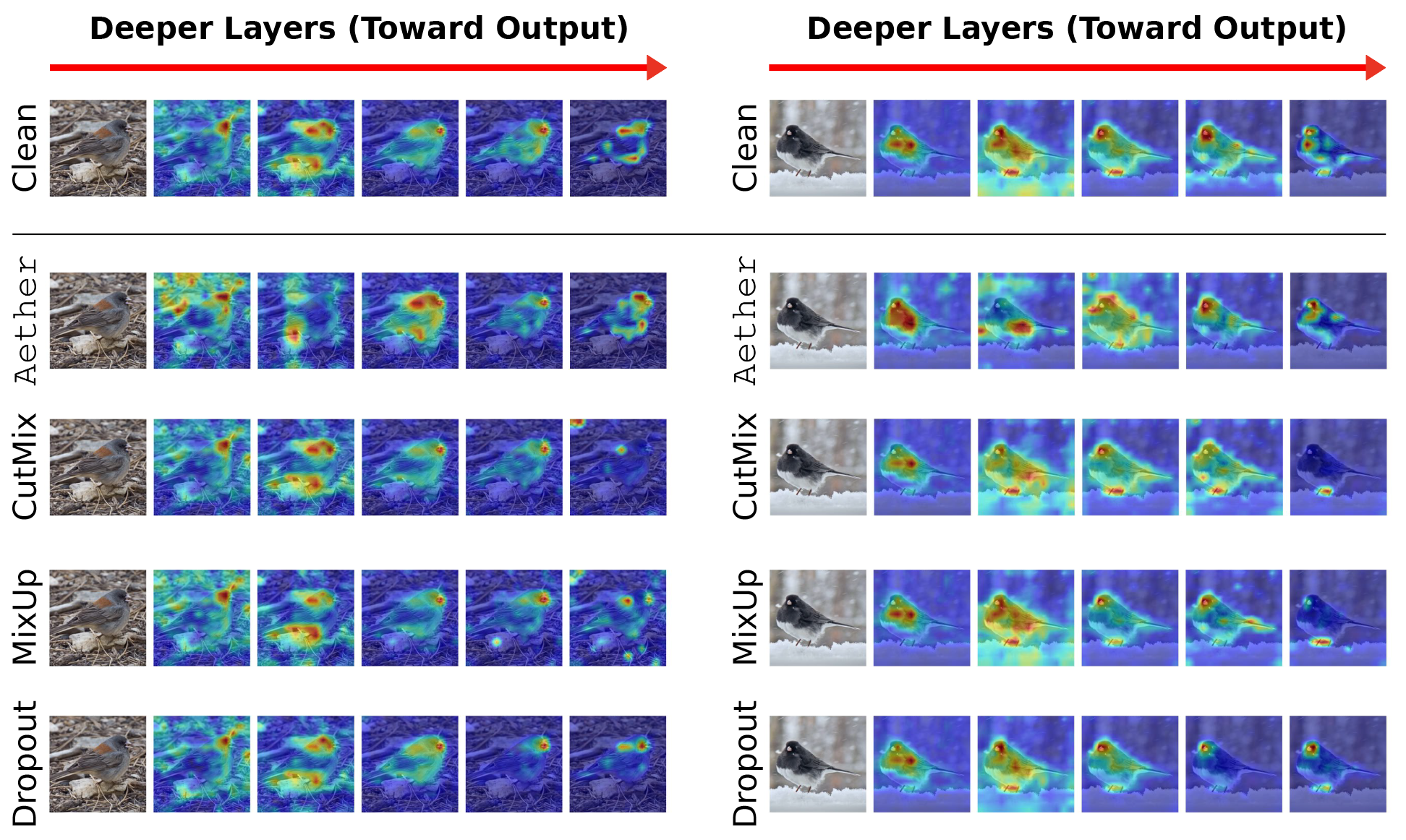}
    %\vspace{-1.2em}
    \caption{Attention visualizations under 20$\times$ perturbations.}
    \label{fig:5_1_denoiser1}
\end{subfigure}
% \hfill
\begin{subfigure}[t]{0.54\linewidth}
    \centering
    \includegraphics[width=\linewidth]{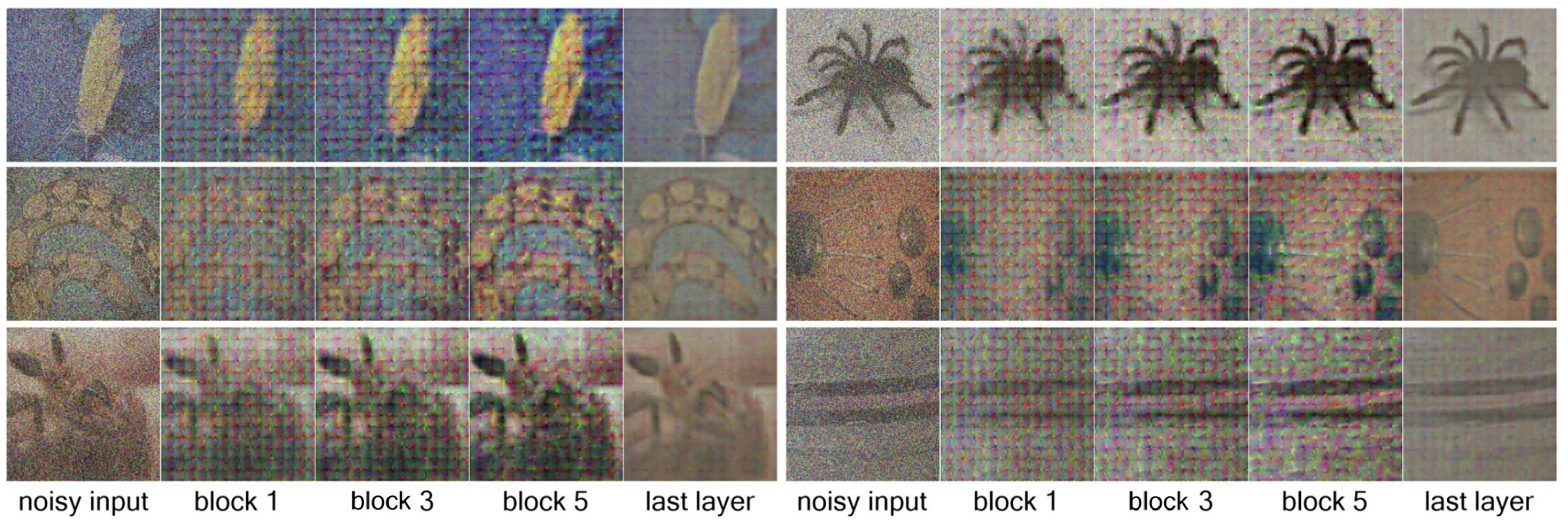}
    %\vspace{-1.2em}
    \caption{Reconstructions under 50$\times$ perturbations.}
    \label{fig:peturb_propagation}
\end{subfigure}
%\vspace{-0.5em}
\caption{\textbf{Robustness under extreme perturbations.} (a) Only \ours\ progressively suppresses perturbations—becoming increasingly similar to the clean image as layers approach the output—yielding class-consistent latents. Conventional augmentations such as CutMix, Mixup, and Dropout retain residual artifacts. (b) Late-layer feature inversions show that \ours\ preserves semantic information even under heavy corruption, producing reconstructions highly similar to the clean image with only minor perturbations.}
\label{fig:robustness}
\vspace{-0.5em}
\end{figure*}

\noindent\textbf{Robustness under extreme perturbations.} We further stress-test robustness by amplifying perturbations far beyond typical levels in the input space. In Fig.~\ref{fig:5_1_denoiser1}, under $20\times$ noise, only \ours recovers attention maps consistent with the clean image. Additionally, feature-map reconstructions under $50\times$ perturbations (Fig.~\ref{fig:peturb_propagation}) preserve semantic structure, demonstrating strong robustness.\\

\begin{table}[t!]
\centering
% \addtolength{\tabcolsep}{-0pt}
\tabcolsep=0.25em
% \tiny
% \scriptsize
\fontsize{8.5}{9}\selectfont
\caption{\textbf{ImageNet-1K top-1 accuracy and relative gains.} 
\ours\ surpasses the performance saturation point of $\mathcal{R}_b$ and consistently improves performance across both vision models and vision language models.}
\vspace{-.5em}
\label{tab:result_vit}

\begin{tabular}{llcc}
\toprule
{Model} & {Augmentation / Setup} & {Top-1 (\%)} & {Gain (\%p)} \\
\midrule
\multicolumn{4}{c}{\textbf{Vision Language Models}}\\
\midrule

\textbf{CLIP} & Baseline & 83.05 & -- \\
& \cellcolor{blue!8}+ \ours & \cellcolor{blue!8}\textbf{84.37} & \cellcolor{blue!8}\textbf{+1.32} \\

\midrule
\textbf{AIMv2} & Baseline & 86.22 & -- \\
& \cellcolor{blue!8}+ \ours & \cellcolor{blue!8}\textbf{87.01} & \cellcolor{blue!8}\textbf{+0.79} \\

\midrule
\textbf{SigLIP 2} & Baseline & 73.64 & -- \\
& \cellcolor{blue!8}+ \ours & \cellcolor{blue!8}\textbf{73.79} & \cellcolor{blue!8}\textbf{+0.15} \\

\midrule
\midrule

\multicolumn{4}{c}{\textbf{Vision Transformer Variants}}\\
\midrule
\textbf{ViT-B} & Baseline & 79.02 & -- \\
& + $\mathcal{R}_b$ {\scriptsize (CutMix + MixUp + DropPath + RandAug)} & 81.17 & +2.15\\
& + $\mathcal{R}_b$ + AugMix & 81.16 & +2.14\\
& + $\mathcal{R}_b$ + RandErase & 81.14 & +2.12\\
& + $\mathcal{R}_b$ + Manifold Mixup & 81.32 & +2.30\\
& + $\mathcal{R}_b$ + Noisy Feature Mixup & 81.28 & +2.26\\
& \cellcolor{blue!8}+ $\mathcal{R}_b$ + \ours & \cellcolor{blue!8}\textbf{82.25} & \cellcolor{blue!8}\textbf{+3.23} \\
& + $\mathcal{R}_b$ + AugMix + RandErase & 81.16 & +2.14\\
& \cellcolor{blue!8}+ $\mathcal{R}_b$ + AugMix + \ours & \cellcolor{blue!8}\textbf{82.43} & \cellcolor{blue!8}\textbf{+3.41} \\
& + $\mathcal{R}_b$ + RandErase + \ours & \textbf{82.38} & \textbf{+3.36} \\
& \cellcolor{blue!8}+ $\mathcal{R}_b$ + AugMix + RandErase + \ours & \cellcolor{blue!8}\textbf{82.51} & \cellcolor{blue!8}\textbf{+3.49} \\

\midrule
\textbf{ViT-S} & Baseline & 77.79 & -- \\
& + $\mathcal{R}_b$ & 78.85 & +1.06\\
& \cellcolor{blue!8}+ $\mathcal{R}_b$ + \ours & \cellcolor{blue!8}\textbf{79.42} & \cellcolor{blue!8}\textbf{+1.63} \\

\midrule
\textbf{ViT-L} & Baseline & 82.24 & -- \\
& + $\mathcal{R}_b$ & 84.71 & +2.47\\
& \cellcolor{blue!8}+ $\mathcal{R}_b$ + \ours & \cellcolor{blue!8}\textbf{85.35} & \cellcolor{blue!8}\textbf{+3.11} \\

\midrule
\textbf{SwinV2} & Baseline & 84.08 & -- \\
& + $\mathcal{R}_b$ & 85.21 & +1.13\\
& \cellcolor{blue!8}+ $\mathcal{R}_b$ + \ours & \cellcolor{blue!8}\textbf{85.30} & \cellcolor{blue!8}\textbf{+1.22} \\

\midrule
\midrule

\textbf{ViT-B} & Baseline & 79.02 & -- \\
& + CutMix~\cite{yun2019cutmix} & 80.08 & +1.34\\
& + MixUp~\cite{zhang2017mixup} & 79.93 & +1.15\\
& + DropPath~\cite{huang2016deep} & 79.65 & +0.80\\
& + RandAug~\cite{cubuk2020randaugment} & 79.64 & +0.79\\
& + AutoAug~\cite{cubuk2019autoaugment} & 79.63 & +0.77\\
& + AugMix~\cite{hendrycks2019augmix} & 79.84 & +1.04\\
& + RandErase~\cite{devries2017improved} & 79.83 & +1.02\\
& \cellcolor{blue!8}+ \ours & \cellcolor{blue!8}\textbf{80.14} & \cellcolor{blue!8}\textbf{+1.41} \\

\midrule
\midrule

\multicolumn{4}{c}{\textbf{CNNs}}\\
\midrule

\textbf{ResNet-50} & Baseline & 79.86 & -- \\
& \cellcolor{blue!8}+ \ours & \cellcolor{blue!8}\textbf{80.04} & \cellcolor{blue!8}\textbf{+0.18} \\

\midrule
\textbf{ResNet-26} & Baseline & 73.20 & -- \\
& \cellcolor{blue!8}+ \ours & \cellcolor{blue!8}\textbf{73.33} & \cellcolor{blue!8}\textbf{+0.13} \\

\bottomrule
\end{tabular}

\vspace{-1em}
\end{table}

\begin{table}[t]
\centering
\tabcolsep=1em
\caption{Cross-modal retrieval on COCO 5K with CLIP ViT-B/16.
\ours is applied only to the vision encoder during fine-tuning, yet improves both
retrieval directions. Recent CLIP-based methods typically gain 0.3\%p -- 0.8\%p.}
\label{tab:retrieval}
\begin{tabular}{lccc}
\toprule
Setting & Baseline (\%)& +\ours (\%) & Gain (\%p) \\
\midrule
COCO I$\rightarrow$T R@1 & 52.1 & 53.4 & +1.3 \\
COCO T$\rightarrow$I R@1 & 32.9 & 33.6 & +0.7 \\
\bottomrule
\end{tabular}
\end{table}

%\begin{figure*}[t]
%\centering
%\includegraphics[width=0.95\linewidth]{images/denoising_effect_new.png}
%\vspace{-1em}
%\caption{\textbf{Reconstructions with \ours-trianed models on intensively perturbed (50$\times$) images.} We highlight the striking capability of \ours-trained models to process heavily corrupted inputs. Late-layer feature inversions show \ours\ preserves semantic information, remaining highly similar to the clean image with only minor perturbations visible, suggesting the model has achieved improved robustness.}
%\label{fig:peturb_propagation}
%\vspace{-.5em}
%\end{figure*}

\noindent\textbf{Implications for fine-grained visual recognition.}
The improved localization and broader attention dynamics directly translate to significant gains in Fine-Grained Visual Classification (FGVC). Tasks in FGVC require the model to discriminate between highly similar subcategories based on subtle, part-level cues. As reported later in the experimental results (Tab.~\ref{tab:downstream_results}), \ours delivers substantial improvements (e.g., CUB accuracy rises from 79.10 to 80.74, and NABirds from 77.87 to 79.52). These gains confirm that by suppressing spurious attention sinks and reinforcing part-level evidence, \ours sharpens the decision boundaries in challenging fine-grained scenarios.

\section{Experiment}
This section performs comprehensive comparisons between baseline models with or without \ours. Our main analysis covers four major axes: (1) vision language models, (2) transformer backbone variants (ViT and Swin), (3) CNNs (ResNet-26/50), and (4) modern self-supervised learning (SSL) frameworks. {Comprehensive ablation studies} are reported in the supplementary material.\\

%\subsection{Implementation Details}
%\noindent\textbf{Models.}
%We use ViTs~\cite{dosovitskiy2020vit} and CNNs~\cite{resnet} for experiments. ViTs' patch size follows the pre-trained architecture, where the ImageNet-21K~\cite{imagenet} variants typically use a patch size of 16 (\eg, ViT-B/16). We also employ SwinV2~\cite{liu2021swin} to evaluate hybrid architectures trained on massive images; ImageNet-21K~\cite{imagenet} pre-trained ViTs to leverage large-scale supervision; diverse SSL-pre-trained (MAE~\cite{mae}, SimMIM~\cite{xie2022simmim}, DiffMAE~\cite{diffmae}, MaskDIT~\cite{maskdit}, and DiffMIM~\cite{diffmim}) ViTs to include diversified training paradigms; and CLIP~\cite{radford2021learning} as a vision–language pre-trained model.

\noindent\textbf{Training setup.} We follow standard fine-tuning protocols from prior work~\cite{vit,deit,mae,touvron2022deit,heo2025maksub, liu2021swin, fini2025multimodal, tschannen2025siglip, radford2021learning, resnet}. Training is conducted using cosine learning rate scheduling~\cite{loshchilov2016sgdr} with the AdamW optimizer~\cite{adamw}.
Regularization and data augmentation include CutMix~\cite{yun2019cutmix}, Mixup~\cite{zhang2017mixup}, DropPath~\cite{huang2016deep}, RandAug~\cite{cubuk2020randaugment}, AutoAug~\cite{cubuk2019autoaugment}, AugMix~\cite{hendrycks2019augmix}, RandErase~\cite{zhong2020random}, \hs{Manifold Mixup~\cite{verma2019manifold}, Noisy Feature Mixup~\cite{lim2021noisy}}, weight decay, and other standard settings to both baseline and \ours-augmented models. Overall, our ImageNet-1K~\cite{imagenet} training is based on the Timm repository~\cite{rw2019timm}. See the supplementary material for detailed experimental settings.\\

%\subsection{Evaluation on Various Architectures}
%We first assess the impact of \ours on standard vision transformer models~\cite{vit}, including ViT-L, ViT-B, and ViT-S, that are pre-trained on ImageNet-1K~\cite{imagenet} and subsequently fine-tuned with/without \ours. Across all ViTs, we evaluate \ours in comparison to a standard augmentation setup $\mathcal{R}_b$, including CutMix~\cite{yun2019cutmix}, MixUp~\cite{zhang2017mixup}, DropPath~\cite{huang2016deep}, and RandAug~\cite{cubuk2020randaugment}, which are commonly used in transformer training pipelines such as timm, as shown in Table~\ref{tab:result_vit}.

\noindent\textbf{On Vision Language Models (VLMs).}
\hs{We evaluate \ours\ on representative VLM architectures, including CLIP~\cite{radford2021learning}, AIMv2~\cite{fini2025multimodal}, and SigLIP 2~\cite{tschannen2025siglip}. For all models, we initialize from the pretrained weights and fine-tune them on ImageNet using their original training setups, preserving the augmentations and regularization strategies used in each model. Our perturbation module is applied only to the vision encoder, leaving the language branch unchanged.}

\hs{For CLIP, the baseline achieves 83.05\%, while applying the standard augmentation recipe $\mathcal{R}_b$ improves performance to 84.10\%. Incorporating \ours\ further raises accuracy to 84.37\%. For AIMv2, \ours\ improves the baseline from 86.22\% to 87.01\%. Similarly, on SigLIP 2, performance increases from 73.64\% to 73.79\%. These results indicate that \ours\ consistently enhances representation robustness and generalizes effectively to large-scale vision language models.}

Beyond ImageNet accuracy, we directly probe cross-modal alignment via image--text
retrieval (Tab.~\ref{tab:retrieval}). Although \ours perturbs only the vision branch,
it improves \emph{both} image-to-text and text-to-image R@1, indicating that a smoother
visual manifold tightens the joint embedding without distorting either modality---
consistent with our isotropy analysis.\\

\noindent\textbf{On ViTs.} $\mathcal{R}_b$ lifts Top-1 from 79.02\% to 81.17\% for ViT-B, while adding other augmentations~\cite{hendrycks2019augmix, devries2017improved} yields no further gains due to overlap along the same three axes (Fig.~\ref{fig:augs_no_complementary}). In contrast, \ours raises accuracy to 82.43\%, expanding the effective regularization space. For ViT-S, we observe a similar trend. The baseline achieves 77.79\%, while $\mathcal{R}_b$ leads to 78.85\%. Incorporating \ours results in 79.42\%, a relative improvement of +2.09\%. For ViT-L, \ours improves performance from the baseline of 82.24\% to 85.35\% when added to the standard augmentation setup. These results indicate that \ours complements standard augmentations. Unlike existing methods, which fall into three primary categories in Fig.~\ref{fig:augs_no_complementary} (b), \ours operates along a distinct regularization axis.\\

\noindent\textbf{On SwinV2-L}~\cite{liu2022swin}, pre-trained with SimMIM~\cite{xie2022simmim} on ImageNet-1K~\cite{imagenet} and then fine-tuned, $\mathcal{R}_b$ reaches 85.21\%; adding \ours nudges it to 85.30\% (+1.45\%), consistent with Swin's stronger built-in locality yet confirming \ours as a complementary regularizer even for locality-aware backbones.\\

%\noindent\textbf{On CLIP}~\cite{radford2021learning} ViT-B, adding \ours lifts accuracy to 84.37\% (+1.59\% over baseline), indicating that \ours transfers beyond vanilla ViTs and remains effective for multi-modal encoders trained with alignment objectives.\\

\noindent\textbf{Extension to CNNs.} \hs{While \ours is primarily evaluated on transformer-based architectures, we also assess its applicability to convolutional networks by applying it to ResNet-50~\cite{resnet} and ResNet-26~\cite{resnet} on ImageNet~\cite{imagenet} classification. \ours improves the top-1 accuracy from 79.86\% to 80.04\% and 73.20\% to 73.33\%, when added to a standard ResNet-50 and ResNet-26 baseline. This confirms that the regularization effect of \ours is not exclusive to transformer models and can extend to CNNs.
We interpret this improvement as further evidence of \ours acting as a general-purpose augmentation method. Unlike traditional augmentations tailored to input-level or region-level transformations, \ours perturbs intermediate features in the embedding space, which also benefits CNN representations by promoting robustness in hidden activations.}
%Nonetheless, the performance gain observed in CNNs is relatively modest compared to the consistent improvements seen in transformers, where attention-based models appear to benefit more from embedding-level regularization. Thus, while \ours is broadly applicable, it is particularly effective in models lacking strong inductive biases---such as ViTs~\cite{vit}---where denoising behavior and semantic localization play a more critical role.

\begin{table*}[t]
\centering
\addtolength{\tabcolsep}{5.2pt}
\small
\caption{\textbf{Top-1 accuracy on ImageNet-1K} of fine-tuning self-supervised pre-trained (SSL) models, with and without \ours. We leverage modern state-of-the-art SSL pre-trained models, including MAE, SimMIM, DiffMAE, MaskDiT, and DiffMIM, and show consistent gains across all the models, demonstrating the superior method-agnostic capability of our method.
}
% \vspace{-.5em}
\label{tab:result_ssl}
%\resizebox{\linewidth}{!}{%
\begin{tabular}{lll c}
\toprule
{SSL Framework} & {Model} & {Method} & {Top-1 Acc (\%)} \\
\midrule
\multirow{6}{*}{\makecell[l]{\textbf{Masked Image} \\ \textbf{Modeling}}} 
& ViT-B~ & MAE~\cite{mae} + $\mathcal{R}_b$ & 82.92 \\
& &  \cellcolor{blue!8} + \ours  & \cellcolor{blue!8} 
 \textbf{83.17} \\
& ViT-L~ & MAE~\cite{mae} + $\mathcal{R}_b$  & 84.42 \\
& & \cellcolor{blue!8} + \ours  & \cellcolor{blue!8} 
 \textbf{84.61} \\
& ViT-B~ & SimMIM~\cite{xie2022simmim} + $\mathcal{R}_b$ & 83.10 \\
& & \cellcolor{blue!8} + \ours & \cellcolor{blue!8} \textbf{83.23} \\
\midrule
\multirow{6}{*}{\makecell[l]{\textbf{Diffusion Model-based} \\ \textbf{Masked Image Modeling}}} 
& ViT-B~ & DiffMAE~\cite{diffmae} + $\mathcal{R}_b$ & 82.18 \\
& & \cellcolor{blue!8} + \ours & \cellcolor{blue!8} 
 \textbf{82.50} \\
& ViT-B~ & MaskDiT~\cite{maskdit} + $\mathcal{R}_b$ & 82.89 \\
& & \cellcolor{blue!8} + \ours & \cellcolor{blue!8} 
 \textbf{83.14} \\
& ViT-B~ & DiffMIM~\cite{diffmim} + $\mathcal{R}_b$ & 83.31 \\
& & \cellcolor{blue!8} + \ours & \cellcolor{blue!8} 
 \textbf{83.52} \\
\bottomrule
\end{tabular}%
% \vspace{-.5em}
\end{table*}

\begin{table*}[t]
\centering
\addtolength{\tabcolsep}{5pt}
\small
\caption{\textbf{Evaluation on downstream tasks.} We evaluate our fine-tuned models using \ours to assess improvements in generalization. We extensively evaluate our method on several datasets, including fine-grained visual classification benchmarks (CUB and NABirds) and large-scale semantic/instance segmentation and object detection benchmarks (ADE20K and COCO). \ours consistently improves performance across all the benchmarks by a large margin. }
\label{tab:downstream_results}
% \vspace{-.5em}
\renewcommand{\arraystretch}{1.2}
\begin{tabular}{lcccc}
\toprule
{Task} & {CUB} & {NABirds} & {ADE20K} & {COCO (AP$^{\text{box}}$/AP$^{\text{mask}}$)} \\
\midrule
Baseline & 79.10 & 77.87 & 43.12 & 46.17 / 40.21 \\
\cellcolor{blue!8} + \ours & \cellcolor{blue!8} \textbf{80.74} & \cellcolor{blue!8} \textbf{79.52} & \cellcolor{blue!8} \textbf{43.56} & \cellcolor{blue!8} \textbf{46.44 / 40.58} \\
\bottomrule
\end{tabular}
% \vspace{-.5em}
\end{table*}

\subsection{Evaluation with SSL-pre-trained Models}
%Self-supervised learning (SSL)-based pre-training~\cite{xie2022simmim, mae} is now central in vision, particularly for training large Vision Transformers effectively. As capacity grows and data efficiency becomes a bottleneck, SSL exploits unlabeled data to learn transferable representations. In practice, a method's ability to integrate with---and improve---SSL pipelines has become a key test of scalability and relevance. 
SSL pre-training~\cite{xie2022simmim, mae} is now fundamental to vision, driving large Vision Transformers with unlabeled data. As capacity scales, effective integration with SSL defines a method's scalability and relevance. Moving beyond supervised pre-trained models in the previous section, we apply \ours at fine-tuning to SSL pre-trained ImageNet-1K checkpoints from MAE (ViT-B/L)~\cite{mae}, SimMIM~\cite{xie2022simmim}, and diffusion-based SSL (DiffMAE, MaskDiT, DiffMIM)~\cite{diffmae,maskdit,diffmim} to test the generalization capacity.

The results in Table~\ref{tab:result_ssl} demonstrate that \ours generalizes effectively to the modern SSL frameworks. Across a variety of SSL-pre-trained models, incorporating \ours during fine-tuning consistently improves performance over established augmentation baselines, even when standard augmentations are already applied, indicating that \ours provides a complementary form of regularization beyond existing methods. This effect is pronounced in diffusion-based SSL, which has gained increasing prominence recently. For instance, \ours lifts DiffMAE~\cite{diffmae} from 82.18\% to 82.50\% and DiffMIM~\cite{diffmim} from 83.31\% to 83.52\%. These improvements, ranging from +0.25\% to +0.34\%, are particularly notable given that they build on already strong SSL baselines. We attribute these gains to reduced pretrain–finetune mismatch: \ours that randomly perturbs embedding, implicitly encourages the model to learn smoother representations, aligning fine-tuning with diffusion-style pre-training. Given the ongoing shift toward diffusion-based pre-training in large vision models, \ours\ provides a new guideline for designing methods that integrate seamlessly with modern SSL approaches and yield consistent complementary gains.

\subsection{Evaluation on Downstream Tasks}
We evaluate \ours on a diverse set of downstream tasks, including fine-grained visual classification (FGVC): CUB~\cite{cub}, NABirds~\cite{nabirds}), semantic segmentation (ADE20K~\cite{ade20k}), and object detection and instance segmentation (COCO~\cite{coco}, as shown in Table~\ref{tab:downstream_results}. \ours improves performance across all tasks, demonstrating its generality beyond image classification. Gains are significant on FGVC datasets: +1.64\%p on CUB~\cite{cub} and +1.65\%p on NABirds~\cite{nabirds}, where subtle part cues (\eg, beak, feather texture) matter and \ours's localization is especially helpful. In semantic segmentation (ADE20K~\cite{ade20k}), where spatially coherent semantic understanding is crucial, mIoU rises from 43.12 to 43.56; in object detection and instance segmentation (COCO~\cite{coco}), AP$^{\text{box}}$ rises from 46.17 to 46.44 and AP$^{\text{mask}}$ rises from 40.21 to 40.58. These improvements suggest that embedding-level perturbation strengthens discriminative features while preserving spatial structure, yielding consistent gains across tasks.

\section{Conclusion}
The conventional paradigm of data augmentation for training vision models, predominantly focused on input-space manipulations and region-level mixing, has reached a saturation point, with the combination yielding diminishing returns even when combined with other strong alternatives.
\hs{Critically, we have shown that these existing recipes are largely ineffective for vision language encoders, as they disrupt the delicate cross-modal alignment between visual tokens and text.}
We have introduced \ours, a simple yet effective plug-in augmentation that \hs{serves as a new axis of regularization: isotropic perturbations directly within the embedding space}.
%could be a breakthrough: controlled perturbations within the embedding space.
Inspired by feature-space regularization in language models and the \hs{diffusion-style} degradation-recovery paradigm in generative pretraining, \ours employs a \hs{variance-preserving} alpha-mixing mechanism to smoothly blend embeddings with isotropic perturbations.

Our analyses have demonstrated that this approach provides a distinct regularization signal \hs{that is uniquely} complementary to the standard recipe \hs{by preserving semantic integrity during the augmentation process}. We validated that \ours enhances localization by promoting broader and more coherent attention dynamics and improves generalization, resulting in flatter loss landscapes. Furthermore, we provided \hs{theoretical and empirical} justification for why embedding-space perturbations succeed where pixel-space approaches falter, highlighting the critical role of maintaining isotropy \hs{to avoid the biased, anisotropic noise typically introduced by input-level transformations}.
%during regularization.
Evaluations across diverse architectures and a wide array of recognition tasks \hs{confirm that} \ours\ consistently improves performance \hs{and robustness, confirming the essential benefit of exploring isotropic embedding-space augmentations for the next generation of visual and multi-modal representation learning.}
%across architectures and tasks, confirming the benefit of exploring embedding-space augmentations.

\section*{Acknowledgments}
This work was supported by the National Research Foundation of Korea (NRF) grants
funded by the Korea government (MSIT) (RS-2025-24803204 and RS-2026-25498839);
by the Institute of Information \& Communications Technology Planning \& Evaluation
(IITP) grant funded by the Korea government (MSIT) (IITP-2026-RS-2026-25546026,
Leading Generative AI Human Resources Development); and by the IITP-ITRC
(Information Technology Research Center) grant funded by the Korea government (MSIT)
(IITP-2026-RS-2020-II201602, 20\%).

% \noindent\textbf{Reproducibility statement.} We conducted all experiments on ImageNet with extra publicly released methods that were all reproducible. Code is available in the Supplementary Material to ensure reproducibility.

%\subsubsection*{Author Contributions}
%If you'd like to, you may include  a section for author contributions as is done
%in many journals. This is optional and at the discretion of the authors.

%\subsubsection*{Acknowledgments}
%Use unnumbered third level headings for the acknowledgments. All
%acknowledgments, including those to funding agencies, go at the end of the paper.

\bibliographystyle{splncs04}
\bibliography{main}
\end{document}